\documentclass[conference]{IEEEtran}
\IEEEoverridecommandlockouts

\usepackage{cite}
\usepackage{amsmath,amssymb,amsfonts}
\usepackage{algorithmic}
\usepackage{graphicx}
\usepackage{textcomp}
\usepackage{xcolor}
\usepackage{subcaption}
\usepackage{hyperref}
\usepackage{makecell}
\usepackage{color}
\usepackage{float}
\def\BibTeX{{\rm B\kern-.05em{\sc i\kern-.025em b}\kern-.08em
    T\kern-.1667em\lower.7ex\hbox{E}\kern-.125emX}}
\begin{document}

\title{SEED: Enhancing Text-to-SQL Performance and Practical Usability Through Automatic Evidence Generation}

\author{\IEEEauthorblockN{Janghyeon Yun}
\IEEEauthorblockA{\textit{Seoul National University}, Korea \\
janghyeon@europa.snu.ac.kr}
\and
\IEEEauthorblockN{Sang-goo Lee\IEEEauthorrefmark{1}\IEEEauthorrefmark{2}}
\IEEEauthorblockA{\IEEEauthorrefmark{1}\textit{Seoul National University}, Korea \\
\IEEEauthorrefmark{2}\textit{IntelliSys}, Korea \\
sglee@europa.snu.ac.kr}
}


\maketitle

\begin{abstract}

Text-to-SQL enables non-experts to retrieve data from databases by converting natural language queries into SQL. However, state-of-the-art text-to-SQL studies rely on the BIRD dataset, which assumes that evidence is provided along with questions. Although BIRD facilitates research advancements, it assumes that users have expertise and domain knowledge, contradicting the fundamental goal of text-to-SQL. In addition, human-generated evidence in BIRD contains defects, including missing or erroneous evidence, which affects model performance.

To address this issue, we propose SEED (System for Evidence Extraction and Domain knowledge generation), an approach that automatically generates evidence to improve performance and practical usability in real-world scenarios. SEED systematically analyzes database schema, description files, and values to extract relevant information. We evaluated SEED on BIRD and Spider, demonstrating that it significantly improves SQL generation accuracy in the no-evidence scenario, and in some cases, even outperforms the setting where BIRD evidence is provided. Our results highlight that SEED-generated evidence not only bridges the gap between research and real-world deployment but also improves the adaptability and robustness of text-to-SQL models. Our code is available at \href{https://github.com/felix01189/SEED}{https://github.com/felix01189/SEED}.

\end{abstract}


\section{Introduction}
Retrieving specific data from a database requires domain knowledge and SQL expertise. However, text-to-SQL alleviates this requirement by translating users' natural language requests into SQL queries, thereby making non-experts easily retrieve data from databases\cite{SURVEY1,SURVEY2,SURVEY3}. Given its potential, extensive research has been conducted on text-to-SQL, and with the rapid advancement of large language models (LLMs), the field has been progressing at an unprecedented pace.

\begin{figure}[t!]
    \vskip\baselineskip
    \begin{subfigure}{0.45\textwidth}
        \centering
        \includegraphics[width=\linewidth]{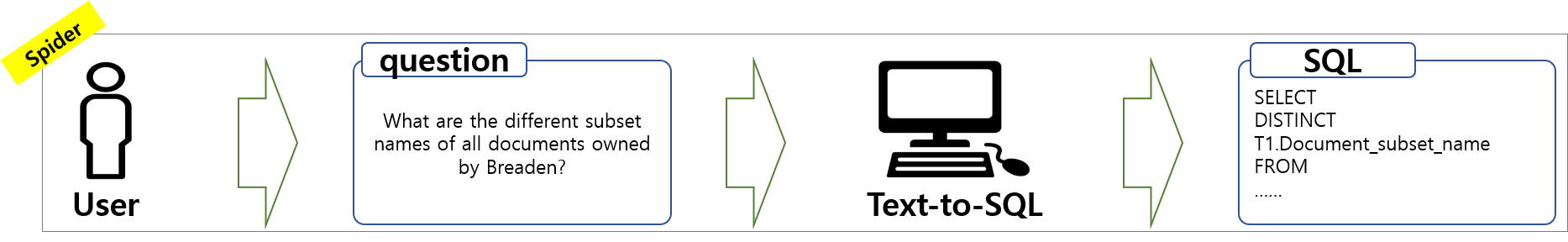}
        \subcaption{Assumption of the text-to-SQL problem in the spider dataset}
    \end{subfigure}
    
    \vskip\baselineskip
    \begin{subfigure}{0.45\textwidth}
        \centering
        \includegraphics[width=\linewidth]{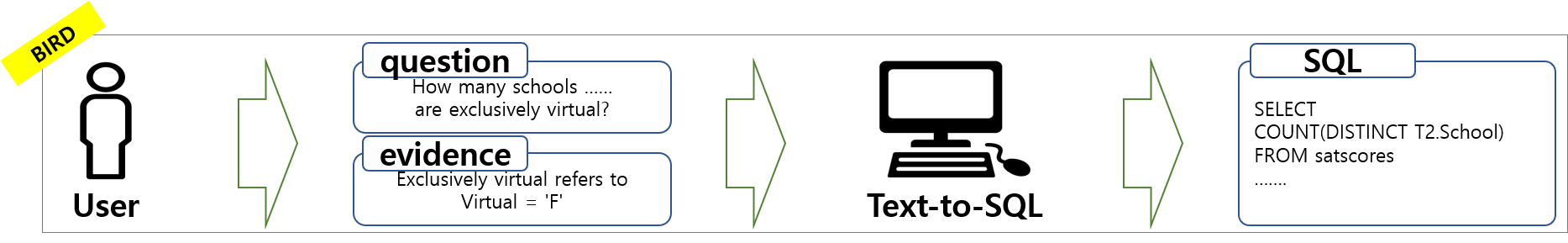}
        \subcaption{Assumption of the text-to-SQL problem in the BIRD dataset}
    \end{subfigure}
    
    \vskip\baselineskip
    \begin{subfigure}{0.45\textwidth}
        \centering
        \includegraphics[width=\linewidth]{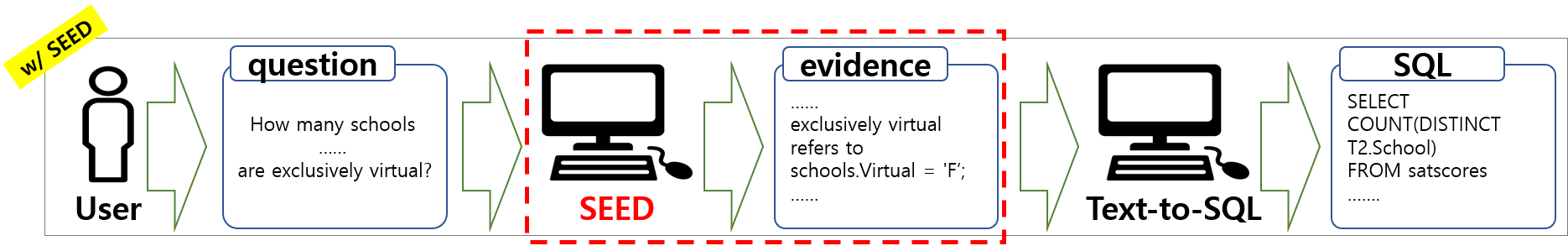}
        \subcaption{Assumption of the text-to-SQL problem with SEED}
    \end{subfigure}
    
    \caption{Diagram of assumptions for the text-to-SQL problems.}
    \label{fig1}
\end{figure}

    
    
    

To validate the effectiveness of these rapidly evolving techniques, numerous researchers have also developed benchmark datasets. Early datasets such as WikiSQL\cite{SEQ2SQL} and Spider\cite{SPIDER} paved the way for text-to-SQL research, and currently the BIRD\cite{BIRD} dataset serves as a prominent benchmark in this domain.

What distinguishes the BIRD\cite{BIRD} dataset from earlier datasets is that it provides information on schema and value through description files and provides the evidence that assists SQL generation. The evidence accompanied by each question includes schema-to-value mappings and mathematical formulas for the calculations necessary for SQL generation. While earlier datasets like Spider\cite{SPIDER} assumed that only questions were provided (as illustrated in Figure \ref{fig1}a), using BIRD evidence assumes that users also provide evidence (as shown in Figure \ref{fig1}b). However, this assumption contradicts the fundamental goal of text-to-SQL.

Despite this unrealistic assumption, most text-to-SQL methods developed based on the BIRD\cite{BIRD}—such as CHASE-SQL\cite{CHASE}, CHESS\cite{CHESS}, and RSL-SQL\cite{RSL}—use these manually provided evidence. In fact, among the top 30 entries on the BIRD leaderboard, all but one unpublished study use evidence\cite{XIYAN,CHASE,DISTILL,CHESS,PURPLE,E-SQL,RSL,MCS,CODES,SENSE,SUPER}. Consequently, applying these state-of-the-art text-to-SQL models in real-world settings, where such evidence is unavailable, creates a gap between academic research and practical deployment, leading to a significant decline in performance. Our experiments confirm that existing text-to-SQL models experience substantial performance degradation when evidence is omitted.

\begin{figure}[t!]
\centerline{\includegraphics[width=\linewidth]{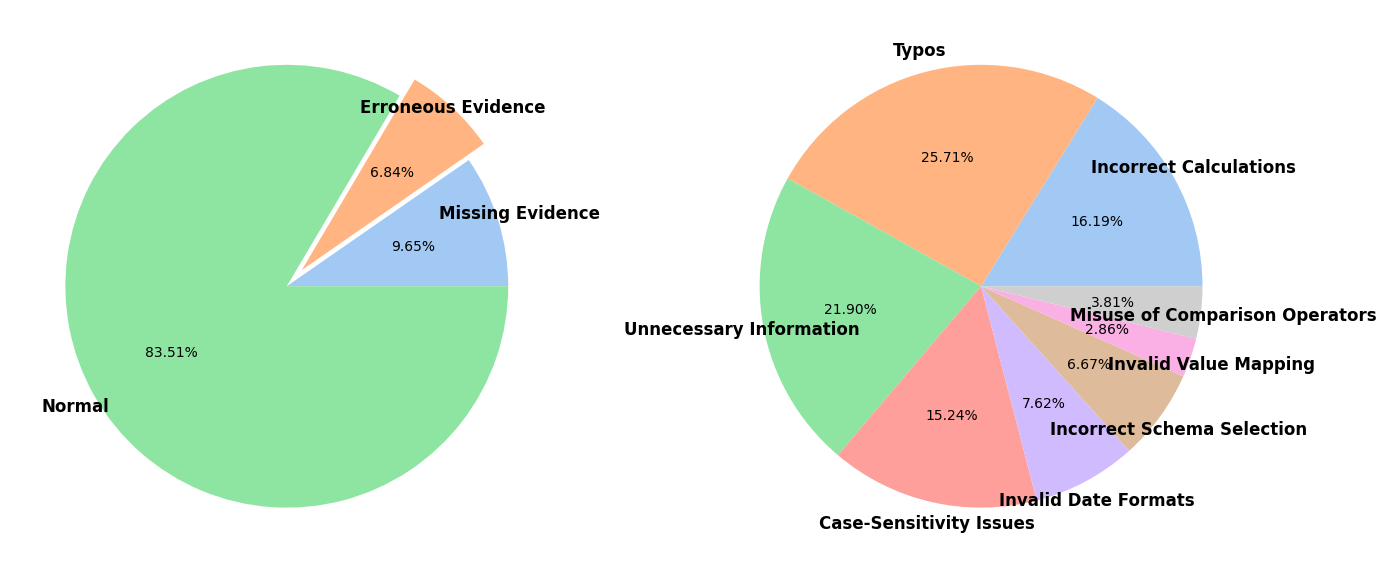}}
\caption{BIRD development set evidence error rate(left) and types(right).}
\label{fig2}
\end{figure}

\begin{table*}[t!]
\centering
\caption{Error samples of BIRD development set evidences}
\normalsize

\begin{tabular}{cl}

\Xhline{2pt}

error type & unnecessary information \\ \hline
question & \makecell[l]{\textcolor{red}{List all the elements} with double bond, consisted in molecule TR024.} \\ \hline
evidence & \makecell[l]{double bond refers to bond\_type = '=';\\
\textcolor{red}{element = 'cl' means Chlorine; element = 'c' means Carbon; element = 'h' means Hydrogen;}\\
\textcolor{red}{element = 'o' means Oxygen, element = 's' means Sulfur; element = 'n' means Nitrogen,}\\
\textcolor{red}{element = 'p' means Phosphorus, element = 'na' means Sodium, element = 'br' means Bromine,}\\
\textcolor{red}{element = 'f' means Fluorine; element = 'i' means Iodine; element = 'sn' means Tin;}\\
\textcolor{red}{element = 'pb' means Lead; element = 'te' means Tellurium; element = 'ca' means Calcium}} \\ \hline
\makecell{revised\\evidence} & \makecell[l]{double bond refers to bond\_type = '=';} \\ \Xhline{2pt}

error type & case-sensitivity issues \\ \hline
question & \makecell[l]{How many cards of legalities whose status is \textcolor{red}{restricted} have text boxes?} \\ \hline
evidence & \makecell[l]{restricted refers to status = \textcolor{red}{'restricted'}; have text boxes refers to is Textless = 0;} \\ \hline
\makecell{revised\\evidence} & \makecell[l]{restricted refers to status = \textcolor{red}{'Restricted'}; have text boxes refers to is Textless = 0;} \\ \Xhline{2pt}

error type & incorrect schema selection \\ \hline
question & \makecell[l]{List down at least five \textcolor{red}{full names}
of superheroes with blue eyes.} \\ \hline
evidence & \makecell[l]{blue eyes refers to colour.colour = 'Blue'\\
WHERE eye\_colour\_id = colour.id;\\
\textcolor{red}{Name} of superheroes refers to \textcolor{red}{superhero\_name};} \\ \hline
\makecell{revised\\evidence} & \makecell[l]{blue eyes refers to colour.colour = 'Blue'\\
WHERE eye\_colour\_id = colour.id;\\
\textcolor{red}{full name} of superheroes refers to \textcolor{red}{full\_name};} \\ \Xhline{2pt}

\end{tabular}
\label{tab1}
\end{table*}

\begin{table}[h!]
    \centering
    \small
    \caption{Performance comparison before and after evidence correction for the 105 erroneous pairs.}
    \begin{tabular}{ccc}
        \hline
        & \multicolumn{2}{c}{EX\%} \\
        \cline{2-3}
        & defective evidence & corrected evidence \\
        \hline
        SFT CodeS-15B & 44.76 & 54.29 \textcolor{blue}{(\textcolor{red}{↑}9.53)} \\
        SFT CodeS-7B & 44.76 & 55.24 \textcolor{blue}{(\textcolor{red}{↑}10.48)}\\
        SFT CodeS-3B & 43.81 & 51.43 \textcolor{blue}{(\textcolor{red}{↑}7.62)}\\
        SFT CodeS-1B & 37.14 & 46.67 \textcolor{blue}{(\textcolor{red}{↑}9.53)}\\
        \hline
    \end{tabular}
    \label{tab2}
\end{table}

Furthermore, our analysis reveals some flaws in the human-generated evidence of BIRD. As Figure \ref{fig2} shows, a thorough review of the development set (1,534 question-SQL-evidence pairs) found that 9.65\% (148 pairs) lacked evidence entirely, while 6.84\% (105 pairs) contained incorrect evidence. The errors in these 105 pairs include incorrect calculations, typos, unnecessary information, case-sensitivity issues, invalid date formats, incorrect schema selection, invalid value mappings, and misuses of comparison operators. Table \ref{tab1} provides examples of these issues. Given that approximately 7\% of the development set contains flawed evidence, these errors introduce noise into text-to-SQL models, potentially limiting their performance. Table \ref{tab2} compares the performance of CodeS\cite{CODES} on the 105 erroneous pairs before and after manually correcting the evidence. This shows that erroneous evidence can significantly degrade performance.

To address these challenges, we propose \textbf{SEED} (\textbf{S}ystem for \textbf{E}vidence \textbf{E}xtraction and \textbf{D}omain knowledge generation). SEED automatically generates evidence by analyzing the schema of the database, the description files and the sampled values. By eliminating reliance on human-generated evidence, SEED aligns with the original objective of text-to-SQL and bridges the gap between academic research and real-world applications.

To validate SEED, we performed experiments using multiple text-to-SQL models under three different conditions: (1) with BIRD evidence, (2) without evidence, and (3) with evidence generated by SEED. Our findings confirm that SEED-generated evidence improves text-to-SQL performance compared to the no-evidence scenario, demonstrating its practical effectiveness.

\begin{table*}[h!]
\centering
\normalsize
\caption{Categories and samples for the evidence of BIRD, and sources of information for each evidence}
\begin{tabular}{cl}

\Xhline{2pt}

\makecell{knowledge type} & Domain Knowledge \\ \hline
question & \makecell[l]{Name the ID and age of patient with two or more laboratory examinations\\
which show their \textcolor{red}{hematoclit level} \textcolor{red}{exceeded the normal range}.} \\ \hline
evidence & \makecell[l]{hematoclit level exceeded the normal range refers to HCT $>$ = 52;} \\ \hline
\makecell{information source} & \makecell[l]{database description file: Laboratory.csv} \\ \hline
information & \makecell[l]{Normal range: 29 $<$ \textcolor{red}{N $<$ 52}} \\ \Xhline{2pt}

\makecell{knowledge type} & Synonym Knowledge \\ \hline
question & \makecell[l]{How many clients opened their accounts in Jesenik branch were \textcolor{red}{women}?} \\ \hline
evidence & \makecell[l]{female refers to gender = 'F'} \\ \hline
\makecell{information source} & \makecell[l]{database description file: client.csv\\or\\database value: select distinct gender from client} \\ \hline
information & \makecell[l]{\textcolor{red}{F:female}\\M:male } \\ \Xhline{2pt}

\makecell{knowledge type} & Value Illustration \\ \hline
question & \makecell[l]{Among the \textcolor{red}{weekly issuance} accounts,
how many have a loan of under 200000?} \\ \hline
evidence & \makecell[l]{frequency = 'POPLATEK TYDNE'\\stands for weekly issuance} \\ \hline
\makecell{information source} & \makecell[l]{database description file: account.csv} \\ \hline
information & \makecell[l]{"POPLATEK MESICNE" stands for monthly issuance\\
\textcolor{red}{"POPLATEK TYDNE" stands for weekly issuance}\\
"POPLATEK PO OBRATU" stands for issuance after transaction} \\ \Xhline{2pt}

\end{tabular}
\label{tab3}
\end{table*}

Our contributions are as follows.
\begin{itemize}
\item Developing an automatic evidence generation system: We introduce SEED, a system that automatically generates evidence to improve text-to-SQL. SEED enhances text-to-SQL applicability in real-world settings where manually generated evidence is unavailable.
\item Bridging research and practical deployment: By automating evidence generation, SEED mitigates the gap between academic studies and real-world implementations, making advanced text-to-SQL models more applicable for real-world adoption.
\item Identifying fundamental issues of using BIRD evidence: We highlight unrealistic assumptions about using BIRD evidence, revealing its inconsistency with the purpose of text-to-SQL. Our analysis confirms that existing text-to-SQL models experience significant performance drops without evidence, highlighting a gap between current research and real-world usability. In addition, we identify errors in the manually generated evidence, emphasizing their negative impact on model performance.
\end{itemize}
Through SEED, our goal is to overcome the limitations of using BIRD evidence and facilitate the broader adoption of text-to-SQL models by making them more robust, practical and effective in real-world scenarios.

\section{Related Work}

\subsection{The BIRD dataset and evidence}

Researchers of the BIRD\cite{BIRD} dataset emphasize that understanding the contents of databases is crucial due to its noisy, messy, and large-scale nature. They argue that external knowledge is necessary to improve text-to-SQL models in understanding database values. They categorize evidence into four types: (1) Numeric reasoning knowledge: Expertise required to perform mathematical calculations. (2) Domain knowledge: Knowledge specific to a particular domain. (3) Synonym knowledge: Information about synonyms, including their meanings and alternative expressions. (4) Value illustration: Detailed descriptions of the values of the database.

However, aside from numeric reasoning knowledge, the remaining three categories—domain knowledge, synonym knowledge, and value illustration—can be derived through a detailed analysis of the database schema, description files, and value samples. Table \ref{tab3} provides examples of these three types of evidence, illustrating that this knowledge can be inferred directly from the given database information.

Thus, the evidence provided in BIRD consists primarily of SQL-related knowledge for mathematical reasoning and domain knowledge extracted through database analysis. This suggests that much of the evidence is not external knowledge, but rather information inherently present within the database itself.

\begin{figure*}[t!]
    \vskip\baselineskip
    \centering
    \begin{subfigure}{0.8\textwidth}
        \centering
        \includegraphics[width=\linewidth]{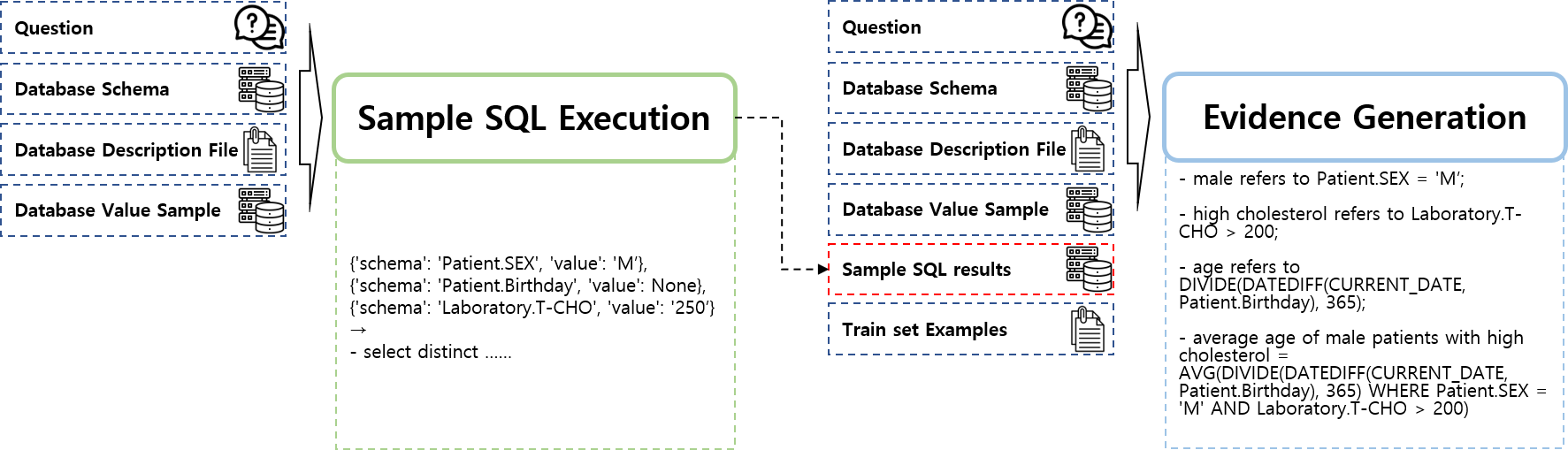}
        \subcaption{The structure of SEED GPT}
    \end{subfigure}
    
    \vskip\baselineskip
    \begin{subfigure}{0.8\textwidth}
        \centering
        \includegraphics[width=\linewidth]{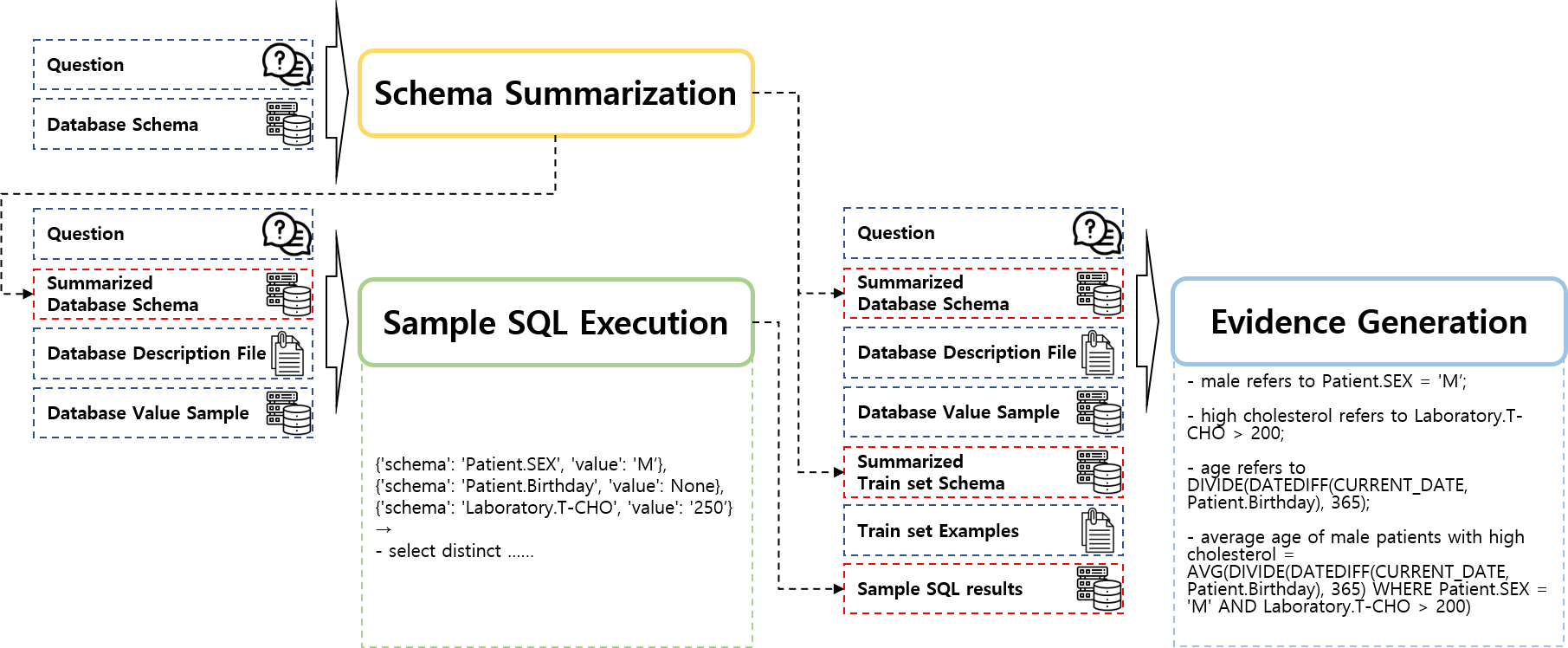}
        \subcaption{The structure of SEED DeepSeek}
    \end{subfigure}
    
    \caption{The structure of SEED}
    \label{fig3}
\end{figure*}

\subsection{Text-to-SQL}

The early text-to-SQL methods were rule-based, relying on predefined patterns\cite{SURVEY2,SURVEY3}. However, these approaches were highly specialized for specific databases, limiting their generalization capabilities. A representative example is NaLIR\cite{NALIR}, which allows users to refine their queries through interactive UI-based modifications.

With the advancement of neural networks, deep learning techniques began to be incorporated into text-to-SQL models\cite{SURVEY1,SURVEY2,SURVEY3}. One of the earliest approaches, Seq2SQL\cite{SEQ2SQL}, utilized a Seq2Seq\cite{SEQ2SEQ} framework to predict appropriate columns and operators based on predefined SQL sketches.

The introduction of Transformer\cite{TRANSFORMER} architectures led to the emergence of pre-trained language models (PLMs) such as BERT\cite{BERT} and T5\cite{T5}, which were subsequently adopted for text-to-SQL tasks\cite{SURVEY1,SURVEY2,SURVEY3}. A notable example is BRIDGE\cite{BRIDGE}, which employs BERT-based encoding combined with a Pointer-Generator Network\cite{PGN} for decoding SQL queries.

Recently, with the rise of closed-source LLMs such as GPT-4\cite{GPT4} and Gemini\cite{GEMINI}, and open-source LLMs like LLaMA\cite{LLAMA} and StarCoder\cite{STARCODER}, most state-of-the-art text-to-SQL methods have adopted LLM-based approaches. C3\cite{C3} is a zero-shot text-to-SQL method built on ChatGPT. DIN-SQL\cite{DIN} decomposes SQL generation into subtasks. DAIL-SQL\cite{DAIL} derives and utilizes effective prompts suitable for text-to-SQL through systematic experiments. MCS-SQL\cite{MCS} generates multiple candidate SQL queries and selects the best one through self-consistency\cite{SELF-CONSISTENCY} mechanisms. XiYan-SQL\cite{XIYAN}, CHASE-SQL\cite{CHASE}, and MSc-SQL\cite{MSC} utilize trained selectors to choose the optimal SQL query from multiple candidates. CHESS\cite{CHESS} incorporates a unit tester to verify SQL predictions. SENSE\cite{SENSE} and CodeS\cite{CODES} focus on fine-tuning smaller models to achieve competitive performance in text-to-SQL tasks. E-SQL\cite{E-SQL} enhances SQL generation by employing a self-polishing\cite{SELF-POLISH} mechanism on the given questions.

As of the time of this paper’s writing, among the top 30 models\cite{XIYAN,CHASE,DISTILL,CHESS,PURPLE,E-SQL,RSL,MCS,CODES,SENSE,SUPER} in the BIRD leaderboard, all but one unpublished method rely on evidence, highlighting its impact in recent text-to-SQL methods.



\section{Methodology}

SEED employs two distinct architectures based on the length of the context input: SEED\(_{\text{gpt}}\) and SEED\(_{\text{deepseek}}\). For scenarios where long context inputs are supported, we use the SEED\(_{\text{gpt}}\) architecture, as depicted in Figure \ref{fig3}a, which processes the entire schema as input. In contrast, in cases with limited context input capacity, we adopt the SEED\(_{\text{deepseek}}\) architecture, as shown in Figure \ref{fig3}b, where the schema is summarized to retain only information relevant to the question.

When GPT-4o is used as the base model, the architecture in Figure \ref{fig3}a is applied. However, for DeepSeek-R1\cite{DEEPSEEK-R1}, which supports a maximum of 8,192 tokens via its API, the architecture in Figure \ref{fig3}b is more appropriate. The SEED framework consists of three key components: Schema Summarization, Sample SQL Execution, and Evidence Generation.

\subsection{Schema Summarization}

The recent study\cite{DISTILL} has highlighted that performing schema pruning as a preprocessing step when leveraging LLMs with strong reasoning capabilities for text-to-SQL tasks can actually degrade SQL generation performance. Building on this insight, SEED does not prune the schema when generating evidence, instead utilizing the entire schema as input to the LLM.

However, there may be cases where the number of input tokens exceeds the limit allowed by the LLM. For instance, the DeepSeek-R1\cite{DEEPSEEK-R1} API imposes a maximum token limit of 8,192. To address such cases, we introduce schema summarization as part of our approach.

Before generating evidence, SEED compares the question with the schema and removes irrelevant information from the schema. This preprocessing step ensures that models with token limitations, such as DeepSeek-R1, can still serve as the base model for SEED.

\subsection{Sample SQL execution}

Consider the process of generating SQL by humans without domain knowledge. When generating SQL without domain knowledge, they can sometimes infer the meaning of keywords in the question by comparing them with the schema. However, in many cases, they need to execute sample SQL queries to inspect database values and fully understand the question's intent.

For example, consider a question that includes the term ``Fremont''. Without executing queries, it is unclear whether ``Fremont'' refers to a county, district, or city. A user would typically run sample queries to check the database before determining the correct column. Similarly, SEED emulates this human process by systematically executing sample SQL queries to generate domain knowledge.

First, SEED extracts keywords that represent database columns and values from the question. Then, it pairs the extracted columns with their corresponding values and generates and executes sample SQL queries for each pair.

The extracted sample data are as follows: unique values are extracted regardless of the data type, and in the case of the string type, similar values are additionally extracted using the LIKE operator and edit distance. The extracted sample data is used to construct the prompt of the next step.

\subsection{Evidence generation}
Once sample SQL results are obtained, SEED generates evidence to assist SQL generation. The evidence generation prompt is structured as follows: an instruction, training set examples, sample SQL results, database schema and question.

To construct effective few-shot examples, SEED identifies similar questions from the training set using similarity-based selection. First, SEED identifies the question most similar to the given query from the training set and then retrieves four more related questions from the same database. We use all-mpnet-base-v2\cite{MPNET} as the embedding model for the comparison of similarity and cosine similarity as the similarity metric.

\begin{table*}[t!]
    \centering
    \scriptsize
    \caption{Performance degradation of text-to-SQL models without evidence on the BIRD dataset and improvement with SEED.}
    \begin{tabular}{cccccccccc}
        \hline
        & \multicolumn{4}{c}{dev EX\%} && \multicolumn{4}{c}{dev VES\%} \\
        \cline{2-5}\cline{7-10}
        & w/o evidence & w/ evidence & w/ SEED\(_{\text{gpt}}\) & w/ SEED\(_{\text{deepseek}}\) && w/o evidence & w/ evidence & w/ SEED\(_{\text{gpt}}\) & w/ SEED\(_{\text{deepseek}}\) \\
        \hline
        \makecell{CHESS\(_{\text{IR+CG+UT}}\)\\(GPT-4o-mini)} & 54.69 & \textbf{63.04} \textcolor{blue}{(\textcolor{red}{↑}8.35)} & \underline{56.26} \textcolor{blue}{(\textcolor{red}{↑}1.57)} & 54.11 \textcolor{blue}{(\textcolor{red}{↓}0.58)} && 56.40 & \textbf{66.64} \textcolor{blue}{(\textcolor{red}{↑}10.24)} & \underline{58.34} \textcolor{blue}{(\textcolor{red}{↑}1.94)} & 55.82 \textcolor{blue}{(\textcolor{red}{↓}0.58)} \\
        
        \makecell{CHESS\(_{\text{IR+SS+CG}}\)\\(GPT-4o-mini)} & 49.61 & \textbf{60.43} \textcolor{blue}{(\textcolor{red}{↑}10.82)} & \underline{54.82} \textcolor{blue}{(\textcolor{red}{↑}5.21)} & 53.65 \textcolor{blue}{(\textcolor{red}{↑}4.04)} && 51.41 & \textbf{64.67} \textcolor{blue}{(\textcolor{red}{↑}13.26)} & \underline{56.75} \textcolor{blue}{(\textcolor{red}{↑}5.34)} & 55.52 \textcolor{blue}{(\textcolor{red}{↑}4.11)} \\
        
        \makecell{RSL-SQL\\(GPT-4o)} & 54.50 & \textbf{65.78} \textcolor{blue}{(\textcolor{red}{↑}11.28)} & \underline{58.28} \textcolor{blue}{(\textcolor{red}{↑}3.78)} & 58.15 \textcolor{blue}{(\textcolor{red}{↑}3.65)} && 56.02 & \textbf{68.31} \textcolor{blue}{(\textcolor{red}{↑}12.29)} & 60.32 \textcolor{blue}{(\textcolor{red}{↑}4.3)} & \underline{64.69} \textcolor{blue}{(\textcolor{red}{↑}8.67)} \\
        
        SFT CodeS-15B & 44.39 & 55.35 \textcolor{blue}{(\textcolor{red}{↑}10.96)} & \underline{56.78} \textcolor{blue}{(\textcolor{red}{↑}12.39)} & \textbf{57.69} \textcolor{blue}{(\textcolor{red}{↑}13.3)} &\makecell{\,\\\,}& 47.22 & 56.84 \textcolor{blue}{(\textcolor{red}{↑}9.62)} & \underline{58.95} \textcolor{blue}{(\textcolor{red}{↑}11.73)} & \textbf{59.33} \textcolor{blue}{(\textcolor{red}{↑}12.11)} \\
        
        SFT CodeS-7B & 41.92 & 54.76 \textcolor{blue}{(\textcolor{red}{↑}12.84)} & \underline{56.52} \textcolor{blue}{(\textcolor{red}{↑}14.60)} & \textbf{56.58} \textcolor{blue}{(\textcolor{red}{↑}14.66)} &\makecell{\,\\\,}& 46.42 & 57.50 \textcolor{blue}{(\textcolor{red}{↑}11.08)} & \textbf{59.65} \textcolor{blue}{(\textcolor{red}{↑}13.23)} & \underline{59.42} \textcolor{blue}{(\textcolor{red}{↑}13.00)} \\
        
        \makecell{DAIL-SQL\\(GPT-4)} & 35.46 & \textbf{56.32} \textcolor{blue}{(\textcolor{red}{↑}20.86)} & 51.63 \textcolor{blue}{(\textcolor{red}{↑}16.17)} & \underline{53.19} \textcolor{blue}{(\textcolor{red}{↑}17.73)} && 36.68 & \textbf{57.70} \textcolor{blue}{(\textcolor{red}{↑}21.02)} & 53.58 \textcolor{blue}{(\textcolor{red}{↑}16.90)} & \underline{54.37} \textcolor{blue}{(\textcolor{red}{↑}17.69)} \\
        \hline
    \end{tabular}
    \label{tab4}
\end{table*}

\begin{table*}[t!]
    \centering
    \small
    \caption{Performance improvement of text-to-SQL models with SEED on the Spider dataset.}
    \begin{tabular}{cccccc}
        \hline
        & \multicolumn{2}{c}{dev EX\%} && \multicolumn{2}{c}{test EX\%} \\
        \cline{2-3}\cline{5-6}
        & w/o SEED & w/ SEED\(_{\text{gpt}}\) && w/o SEED & w/ SEED\(_{\text{gpt}}\) \\
        \hline
        SFT CodeS-15B & 85.6 & 87.3 \textcolor{blue}{(\textcolor{red}{↑}1.7)} && 85.0 & 86.4 \textcolor{blue}{(\textcolor{red}{↑}1.4)} \\
        SFT CodeS-7B & 86.4 & 86.8 \textcolor{blue}{(\textcolor{red}{↑}0.4)} && 84.7 & 86.1 \textcolor{blue}{(\textcolor{red}{↑}1.4)} \\
        C3 (ChatGPT) & 82.0 & 86.6 \textcolor{blue}{(\textcolor{red}{↑}4.6)} && 80.1 & 84.0 \textcolor{blue}{(\textcolor{red}{↑}3.9)} \\
        \hline
    \end{tabular}
    \label{tab5}
\end{table*}

\section{Experiments}

\subsection{Datasets}

The \textbf{BIRD}\cite{BIRD} dataset bridges the gap between text-to-SQL research and real-world applications by incorporating noisy, large-scale data. It consists of 12,751 text-to-SQL pairs across 95 databases (33.4 GB) and 37 domains, uniquely providing database description files and evidence.

The \textbf{Spider}\cite{SPIDER} dataset includes 200 databases, 10,181 questions, and 5,693 complex SQL queries featuring advanced patterns like JOIN, GROUP BY, and EXISTS. Created by 11 students over 1,000 hours, it ensures diverse, multi-table queries for better generalization.

\subsection{Evaluation metrics}

Evaluating text-to-SQL models based on exact SQL matches can lead to false negatives, as different queries may be semantically equivalent. To address this, BIRD and Spider use the \textbf{execution accuracy (EX)}, which compares execution results instead of syntax. In addition, BIRD introduces the \textbf{valid efficiency score (VES)}, which extends EX by factoring in execution time, rewarding more efficient queries with higher scores.

\subsection{Baseline Methods}

To evaluate the effectiveness of SEED-generated evidence, we selected state-of-the-art text-to-SQL models from the BIRD and Spider leaderboards that had publicly available implementations at the time of writing. The chosen models include CHESS\cite{CHESS} and RSL-SQL\cite{RSL}, which represent the latest methods on the leaderboard; CodeS\cite{CODES}, a fine-tuned text-to-SQL model; DAIL-SQL\cite{DAIL} and C3\cite{C3}, which exemplify in-context learning (ICL) approaches.

\subsubsection{CHESS} The CHESS\cite{CHESS} framework approaches text-to-SQL as a multi-agent system comprising four key components: information retriever (IR) – retrieves relevant database values and descriptions, schema selector (SS) – filters out unnecessary schema elements, candidate generator (CG) – generates multiple SQL candidates, unit tester (UT) – performs unit testing to evaluate candidate SQL queries. Additionally, CHESS provides guidelines for configuring these agents based on constraints such as computational budgets, making it a versatile framework for text-to-SQL tasks.

\subsubsection{RSL-SQL} Recent study\cite{DISTILL} has revealed that while schema linking is commonly employed to reduce noise and computational overhead, it can introduce potential risks. To mitigate these drawbacks, the researchers of RSL-SQL\cite{RSL} proposed a bidirectional schema linking approach. First, the full schema is linked and used to generate a preliminary SQL query. Then, the schema elements referenced in the query are extracted. By combining forward and backward schema linking, RSL-SQL achieves a robust and effective schema linking process.

\subsubsection{CodeS} CodeS\cite{CODES} addresses key challenges in text-to-SQL research, such as reliance on closed-source LLMs (e.g., GPT-4 and Gemini), which raise privacy concerns and incur high API costs. To overcome these issues, CodeS fine-tunes StarCoder\cite{STARCODER} to better suit text-to-SQL tasks. The model integrates the schema linking method of RESDSQL\cite{RESDSQL} and enhances database value referencing through a combination of the BM25 index and the longest common substring method. Despite its relatively compact size of up to 15 billion parameters, CodeS outperforms DIN-SQL\cite{DIN}, which uses GPT-4, demonstrating its efficiency and effectiveness.

\subsubsection{DAIL-SQL} With the rise of text-to-SQL in-context learning (ICL), the DAIL-SQL\cite{DAIL} research team highlighted the importance of systematic prompt engineering. Their study explored key aspects, including how to format database schema in prompts, retrieve effective few-shot examples, and represent examples within prompts. By optimizing these factors, DAIL-SQL achieved superior SQL generation performance through a carefully designed prompt strategy.

\subsubsection{C3} C3\cite{C3} is a zero-shot text-to-SQL approach developed to address the limitations of few-shot methods, which often require more than 10,000 tokens, as well as the lower performance of zero-shot models compared to fine-tuned alternatives. The model comprises three stages: Clear Prompting (CP), which establishes schema linking through zero-shot prompt instructions; Calibration with Hints (CH), which identifies biases in ChatGPT—such as over-selecting columns or retrieving excessive values—through error analysis and mitigates these biases by providing specific hints (e.g., “use COUNT(*), LEFT JOIN, or OR only when necessary”); and Consistent Output (CO), which reduces the inherent randomness of LLMs by executing multiple runs and applying a voting mechanism.

\subsection{Implementation details}

SEED\(_{\text{gpt}}\) consists of two stages: sample SQL execution and evidence generation, without schema summarization. The sample SQL execution stage uses gpt-4o-mini, while the evidence generation stage uses gpt-4o.

On the other hand, SEED\(_{\text{deepseek}}\) performs schema summarization twice: once for the database corresponding to the question and once for the train set examples. SEED\(_{\text{deepseek}}\) uses DeepSeek-R1\cite{DEEPSEEK-R1} as the base model for all stages.

\subsection{Results}

\subsubsection{BIRD}

The Table \ref{tab4} presents the performance of SQL generation in BIRD in three different settings: no evidence provided, human-annotated evidence (from BIRD) provided, and SEED-generated evidence provided. 
The first key observation is that without human-generated evidence significantly degrades performance across all models. In particular, the largest performance gap in EX is observed in DAIL-SQL\cite{DAIL}, with a difference of 20.86\%. Even the model with the smallest gap still showed a performance decrease of 8.35\% when evidence was excluded. This highlights the substantial disparity between research settings and real-world applications, where the absence of evidence can severely degrade the performance of SQL generation.

Next, when comparing SEED-generated evidence with the no-evidence condition, we confirm improvements in EX and VES across most models, with some cases even surpassing performance in the BIRD evidence setting. We observe an improvement of up to 17.73\% in EX and up to 17.69\% in VES, indicating that SEED not only improves accuracy but also generates more efficient queries. These results confirm that SEED-generated evidence helps bridge the research-to-reality gap and enhances the practical usability of text-to-SQL models.

\begin{table*}[h!]
\centering
\normalsize
\caption{Examples of BIRD evidence and SEED\(_{\text{deepseek}}\) evidence and revised evidence}
\begin{tabular}{cl}

\Xhline{2pt}

BIRD evidence & \makecell[l]{Magnet schools or offer a magnet program means that Magnet = 1} \\ \hline
SEED\(_{\text{deepseek}}\) & \makecell[l]{SAT test takers of over 500 refers to `satscores`.`NumTstTakr` $>$ 500;\\magnet schools or offer a magnet program refers to `schools`.`Magnet` = 1;\\\textcolor{red}{join on `satscores`.`cds` = `schools`.`CDSCode`}} \\ \hline
SEED\(_{\text{revised}}\) & \makecell[l]{SAT test takers of over 500 refers to `satscores`.`NumTstTakr` $>$ 500;\\magnet schools or offer a magnet program refers to `schools`.`Magnet` = 1}
\\ \Xhline{2pt}

\end{tabular}
\label{tab6}
\end{table*}

\begin{table*}[t!]
    \centering
    \caption{Performance improvement of text-to-SQL models with SEED on the Spider dataset.}
    \begin{tabular}{cccccccc}
        \hline
        & \multicolumn{3}{c}{dev EX\%} && \multicolumn{3}{c}{dev VES\%} \\
        \cline{2-4}\cline{6-8}
        & w/o SEED & w/ SEED\(_{\text{deepseek}}\) & w/ SEED\(_{\text{revised}}\) && w/o SEED & w/ SEED\(_{\text{deepseek}}\) & w/ SEED\(_{\text{revised}}\) \\
        \hline
        CHESS\(_{\text{IR+CG+UT}}\) & 54.69 & 54.11 \textcolor{blue}{(\textcolor{red}{↓}0.58)} & 55.48 \textcolor{blue}{(\textcolor{red}{↑}0.79)} && 56.40 & 55.82 \textcolor{blue}{(\textcolor{red}{↓}0.58)} & 57.39 \textcolor{blue}{(\textcolor{red}{↑}0.99)} \\
        SFT CodeS-15B & 44.39 & 57.69 \textcolor{blue}{(\textcolor{red}{↑}13.30)} & 56.39 \textcolor{blue}{(\textcolor{red}{↑}12.00)} && 47.22 & 59.33 \textcolor{blue}{(\textcolor{red}{↑}12.11)} & 58.44 \textcolor{blue}{(\textcolor{red}{↑}11.22)} \\
        SFT CodeS-7B & 41.92 & 56.58 \textcolor{blue}{(\textcolor{red}{↑}14.66)} & 55.80 \textcolor{blue}{(\textcolor{red}{↑}13.88)} && 46.42 & 59.42 \textcolor{blue}{(\textcolor{red}{↑}13.00)} & 58.42 \textcolor{blue}{(\textcolor{red}{↑}12.00)} \\
        \hline
    \end{tabular}
    \label{tab7}
\end{table*}

\subsubsection{Performance degradation analysis}

However, we note that for CHESS\(_{\text{IR+CG+UT}}\)\cite{CHESS}, SEED\(_{\text{deepseek}}\) performed slightly worse than the no-evidence condition. For the CodeS\cite{CODES} model, the evidence generated by SEED\(_{\text{deepseek}}\) outperformed that of SEED\(_{\text{gpt}}\). However, for the CHESS\(_{\text{IR+CG+UT}}\) model, using SEED\(_{\text{deepseek}}\) evidence resulted in lower performance compared to not using evidence at all. To investigate the cause of this discrepancy, we analyzed both the CHESS model and the evidence generated by SEED, leading to the following observations: (1) SEED utilized human-generated BIRD evidence as few-shot examples. However, it produced information that was not present in the examples or slightly altered the format of the evidence. In particular, as shown in Table \ref{tab6}, the most significant difference was that SEED provided additional information about joins. (2) Earlier studies, such as CodeS and DAIL-SQL\cite{DAIL}, adopted a straightforward approach by simply concatenating the evidence with the question. In contrast, more recent models like CHESS actively incorporate evidence multiple times within each agent, and the prompt for CHESS not only includes direct guidelines on how to utilize evidence but also explicitly specifies the type of information contained in the evidence.

Based on these observations, we hypothesized that recent models like CHESS are optimized for the format of human-generated BIRD evidence through prompt engineering. To validate this hypothesis, we revised SEED evidence by removing join-related information, its most significant difference, using DeepSeek-V3\cite{DEEPSEEK}. We then evaluated the performance of CHESS and CodeS using this revised SEED (SEED\(_{\text{revised}}\)) evidence. As shown in Table \ref{tab7}, CHESS\(_{\text{IR+CG+UT}}\) exhibited an improvement of 1.37\% in EX when using SEED\(_{\text{revised}}\) compared to SEED\(_{\text{deepseek}}\), while CodeS experienced a performance decrease of 1.3\%. This confirms our hypothesis that modifying the SEED evidence to resemble human-generated evidence improves CHESS\(_{\text{IR+CG+UT}}\) performance while reducing CodeS performance. These findings highlight the need for future research on optimizing evidence formats based on how models utilize evidence.

\subsubsection{Spider}

To further validate the robustness of SEED, we performed experiments on the Spider dataset. We compared two scenarios: no evidence provided, SEED-generated evidence provided. For evaluation, we selected one fine-tuned model (CodeS\cite{CODES}) and one ICL-based model (C3\cite{C3}). Since Spider does not have database description files, we generated them using DeepSeek-V3\cite{DEEPSEEK}. The results in Table \ref{tab5} show that the performance of all models has increased, confirming that the evidence generated by SEED improves SQL generation in different datasets and methodologies.

\section{Conclusion}

In this study, we confirmed that the reliance on human-curated evidence creates a gap between academic research and real-world applications. To address this limitation, we proposed SEED, a system that autonomously generates evidence without human intervention. Through experiments, we demonstrated that SEED significantly enhances the performance of text-to-SQL methods in a realistic scenario where evidence is not available.

By bridging the gap between academic research and practical deployment in text-to-SQL systems, our work contributes to making existing and future methodologies more applicable in real-world scenarios. We believe that our findings will guide future research towards more practical and scalable solutions, ultimately facilitating the broader adoption of text-to-SQL in real-world applications.

\bibliographystyle{IEEEtran} 
\bibliography{bio.bib}

\begin{thebibliography}{10}
\providecommand{\url}[1]{#1}
\csname url@samestyle\endcsname
\providecommand{\newblock}{\relax}
\providecommand{\bibinfo}[2]{#2}
\providecommand{\BIBentrySTDinterwordspacing}{\spaceskip=0pt\relax}
\providecommand{\BIBentryALTinterwordstretchfactor}{4}
\providecommand{\BIBentryALTinterwordspacing}{\spaceskip=\fontdimen2\font plus
\BIBentryALTinterwordstretchfactor\fontdimen3\font minus \fontdimen4\font\relax}
\providecommand{\BIBforeignlanguage}[2]{{%
\expandafter\ifx\csname l@#1\endcsname\relax
\typeout{** WARNING: IEEEtran.bst: No hyphenation pattern has been}%
\typeout{** loaded for the language `#1'. Using the pattern for}%
\typeout{** the default language instead.}%
\else
\language=\csname l@#1\endcsname
\fi
#2}}
\providecommand{\BIBdecl}{\relax}
\BIBdecl

\bibitem{SURVEY1}
\BIBentryALTinterwordspacing
B.~Qin, B.~Hui, L.~Wang, M.~Yang, J.~Li, B.~Li, R.~Geng, R.~Cao, J.~Sun, L.~Si, F.~Huang, and Y.~Li, ``A survey on text-to-sql parsing: Concepts, methods, and future directions,'' 2022. [Online]. Available: \url{https://arxiv.org/abs/2208.13629}
\BIBentrySTDinterwordspacing

\bibitem{SURVEY2}
\BIBentryALTinterwordspacing
L.~Shi, Z.~Tang, N.~Zhang, X.~Zhang, and Z.~Yang, ``A survey on employing large language models for text-to-sql tasks,'' 2024. [Online]. Available: \url{https://arxiv.org/abs/2407.15186}
\BIBentrySTDinterwordspacing

\bibitem{SURVEY3}
\BIBentryALTinterwordspacing
Z.~Hong, Z.~Yuan, Q.~Zhang, H.~Chen, J.~Dong, F.~Huang, and X.~Huang, ``Next-generation database interfaces: A survey of llm-based text-to-sql,'' 2024. [Online]. Available: \url{https://arxiv.org/abs/2406.08426}
\BIBentrySTDinterwordspacing

\bibitem{SEQ2SQL}
\BIBentryALTinterwordspacing
V.~Zhong, C.~Xiong, and R.~Socher, ``Seq2sql: Generating structured queries from natural language using reinforcement learning,'' 2017. [Online]. Available: \url{https://arxiv.org/abs/1709.00103}
\BIBentrySTDinterwordspacing

\bibitem{SPIDER}
T.~Yu, R.~Zhang, K.~Yang, M.~Yasunaga, D.~Wang, Z.~Li, J.~Ma, I.~Li, Q.~Yao, S.~Roman, Z.~Zhang, and D.~R. Radev, ``Spider: A large-scale human-labeled dataset for complex and cross-domain semantic parsing and text-to-sql task,'' in \emph{2018 CONFERENCE ON EMPIRICAL METHODS IN NATURAL LANGUAGE PROCESSING (EMNLP 2018)}.\hskip 1em plus 0.5em minus 0.4em\relax Google; Facebook; Bloomberg; Salesforce; Apple; Amazon; Baidu; Grammarly; Naver Labs Europe; FWO; KU Leuven, Dept Comp Sci; CVTE; Ebay; Microsoft; Naver Line; Oracle; Polya; Huawei; Duolingo; Figure Eight; Nuance, 2018, pp. 3911--3921, conference on Empirical Methods in Natural Language Processing (EMNLP), Brussels, BELGIUM, OCT 31-NOV 04, 2018.

\bibitem{BIRD}
\BIBentryALTinterwordspacing
J.~Li, B.~Hui, G.~Qu, J.~Yang, B.~Li, B.~Li, B.~Wang, B.~Qin, R.~Geng, N.~Huo, X.~Zhou, M.~Chenhao, G.~Li, K.~Chang, F.~Huang, R.~Cheng, and Y.~Li, ``Can llm already serve as a database interface? a big bench for large-scale database grounded text-to-sqls,'' in \emph{Advances in Neural Information Processing Systems}, A.~Oh, T.~Naumann, A.~Globerson, K.~Saenko, M.~Hardt, and S.~Levine, Eds., vol.~36.\hskip 1em plus 0.5em minus 0.4em\relax Curran Associates, Inc., 2023, pp. 42\,330--42\,357. [Online]. Available: \url{https://proceedings.neurips.cc/paper_files/paper/2023/file/83fc8fab1710363050bbd1d4b8cc0021-Paper-Datasets_and_Benchmarks.pdf}
\BIBentrySTDinterwordspacing

\bibitem{CHASE}
\BIBentryALTinterwordspacing
M.~Pourreza, H.~Li, R.~Sun, Y.~Chung, S.~Talaei, G.~T. Kakkar, Y.~Gan, A.~Saberi, F.~Ozcan, and S.~O. Arik, ``Chase-sql: Multi-path reasoning and preference optimized candidate selection in text-to-sql,'' 2024. [Online]. Available: \url{https://arxiv.org/abs/2410.01943}
\BIBentrySTDinterwordspacing

\bibitem{CHESS}
\BIBentryALTinterwordspacing
S.~Talaei, M.~Pourreza, Y.-C. Chang, A.~Mirhoseini, and A.~Saberi, ``Chess: Contextual harnessing for efficient sql synthesis,'' 2024. [Online]. Available: \url{https://arxiv.org/abs/2405.16755}
\BIBentrySTDinterwordspacing

\bibitem{RSL}
\BIBentryALTinterwordspacing
Z.~Cao, Y.~Zheng, Z.~Fan, X.~Zhang, W.~Chen, and X.~Bai, ``Rsl-sql: Robust schema linking in text-to-sql generation,'' 2024. [Online]. Available: \url{https://arxiv.org/abs/2411.00073}
\BIBentrySTDinterwordspacing

\bibitem{XIYAN}
\BIBentryALTinterwordspacing
Y.~Gao, Y.~Liu, X.~Li, X.~Shi, Y.~Zhu, Y.~Wang, S.~Li, W.~Li, Y.~Hong, Z.~Luo, J.~Gao, L.~Mou, and Y.~Li, ``Xiyan-sql: A multi-generator ensemble framework for text-to-sql,'' 2024. [Online]. Available: \url{https://arxiv.org/abs/2411.08599}
\BIBentrySTDinterwordspacing

\bibitem{DISTILL}
\BIBentryALTinterwordspacing
K.~Maamari, F.~Abubaker, D.~Jaroslawicz, and A.~Mhedhbi, ``The death of schema linking? text-to-sql in the age of well-reasoned language models,'' 2024. [Online]. Available: \url{https://arxiv.org/abs/2408.07702}
\BIBentrySTDinterwordspacing

\bibitem{PURPLE}
T.~Ren, Y.~Fan, Z.~He, R.~Huang, J.~Dai, C.~Huang, Y.~Jing, K.~Zhang, Y.~Yang, and X.~S. Wang, ``Purple: Making a large language model a better sql writer,'' in \emph{2024 IEEE 40th International Conference on Data Engineering (ICDE)}, 2024, pp. 15--28.

\bibitem{E-SQL}
\BIBentryALTinterwordspacing
H.~A. Caferoğlu and Özgür Ulusoy, ``E-sql: Direct schema linking via question enrichment in text-to-sql,'' 2024. [Online]. Available: \url{https://arxiv.org/abs/2409.16751}
\BIBentrySTDinterwordspacing

\bibitem{MCS}
\BIBentryALTinterwordspacing
D.~Lee, C.~Park, J.~Kim, and H.~Park, ``Mcs-sql: Leveraging multiple prompts and multiple-choice selection for text-to-sql generation,'' 2024. [Online]. Available: \url{https://arxiv.org/abs/2405.07467}
\BIBentrySTDinterwordspacing

\bibitem{CODES}
\BIBentryALTinterwordspacing
H.~Li, J.~Zhang, H.~Liu, J.~Fan, X.~Zhang, J.~Zhu, R.~Wei, H.~Pan, C.~Li, and H.~Chen, ``Codes: Towards building open-source language models for text-to-sql,'' \emph{Proc. ACM Manag. Data}, vol.~2, no.~3, May 2024. [Online]. Available: \url{https://doi.org/10.1145/3654930}
\BIBentrySTDinterwordspacing

\bibitem{SENSE}
\BIBentryALTinterwordspacing
J.~Yang, B.~Hui, M.~Yang, J.~Yang, J.~Lin, and C.~Zhou, ``Synthesizing text-to-{SQL} data from weak and strong {LLM}s,'' in \emph{Proceedings of the 62nd Annual Meeting of the Association for Computational Linguistics (Volume 1: Long Papers)}, L.-W. Ku, A.~Martins, and V.~Srikumar, Eds.\hskip 1em plus 0.5em minus 0.4em\relax Bangkok, Thailand: Association for Computational Linguistics, Aug. 2024, pp. 7864--7875. [Online]. Available: \url{https://aclanthology.org/2024.acl-long.425/}
\BIBentrySTDinterwordspacing

\bibitem{SUPER}
\BIBentryALTinterwordspacing
B.~Li, Y.~Luo, C.~Chai, G.~Li, and N.~Tang, ``The dawn of natural language to sql: Are we fully ready?'' \emph{Proc. VLDB Endow.}, vol.~17, no.~11, p. 3318–3331, Aug. 2024. [Online]. Available: \url{https://doi.org/10.14778/3681954.3682003}
\BIBentrySTDinterwordspacing

\bibitem{NALIR}
\BIBentryALTinterwordspacing
F.~Li and H.~V. Jagadish, ``Nalir: an interactive natural language interface for querying relational databases,'' in \emph{Proceedings of the 2014 ACM SIGMOD International Conference on Management of Data}, ser. SIGMOD '14.\hskip 1em plus 0.5em minus 0.4em\relax New York, NY, USA: Association for Computing Machinery, 2014, p. 709–712. [Online]. Available: \url{https://doi.org/10.1145/2588555.2594519}
\BIBentrySTDinterwordspacing

\bibitem{SEQ2SEQ}
I.~Sutskever, ``Sequence to sequence learning with neural networks,'' \emph{arXiv preprint arXiv:1409.3215}, 2014.

\bibitem{TRANSFORMER}
A.~Vaswani, ``Attention is all you need,'' \emph{Advances in Neural Information Processing Systems}, 2017.

\bibitem{BERT}
Kenton, J.~Devlin, M.-W. Chang, Toutanova, and L.~Kristina, ``Bert: Pre-training of deep bidirectional transformers for language understanding,'' in \emph{Proceedings of naacL-HLT}, vol.~1, no.~2.\hskip 1em plus 0.5em minus 0.4em\relax Minneapolis, Minnesota, 2019.

\bibitem{T5}
Raffel, Colin, Shazeer, Noam, Roberts, and e.~a. Adam, ``Exploring the limits of transfer learning with a unified text-to-text transformer,'' \emph{Journal of machine learning research}, vol.~21, no. 140, pp. 1--67, 2020.

\bibitem{BRIDGE}
\BIBentryALTinterwordspacing
X.~V. Lin, R.~Socher, and C.~Xiong, ``Bridging textual and tabular data for cross-domain text-to-{SQL} semantic parsing,'' in \emph{Findings of the Association for Computational Linguistics: EMNLP 2020}, T.~Cohn, Y.~He, and Y.~Liu, Eds.\hskip 1em plus 0.5em minus 0.4em\relax Online: Association for Computational Linguistics, Nov. 2020, pp. 4870--4888. [Online]. Available: \url{https://aclanthology.org/2020.findings-emnlp.438/}
\BIBentrySTDinterwordspacing

\bibitem{PGN}
\BIBentryALTinterwordspacing
A.~See, P.~J. Liu, and C.~D. Manning, ``Get to the point: Summarization with pointer-generator networks,'' in \emph{Proceedings of the 55th Annual Meeting of the Association for Computational Linguistics (Volume 1: Long Papers)}, R.~Barzilay and M.-Y. Kan, Eds.\hskip 1em plus 0.5em minus 0.4em\relax Vancouver, Canada: Association for Computational Linguistics, Jul. 2017, pp. 1073--1083. [Online]. Available: \url{https://aclanthology.org/P17-1099/}
\BIBentrySTDinterwordspacing

\bibitem{GPT4}
\BIBentryALTinterwordspacing
OpenAI, J.~Achiam, S.~Adler, S.~Agarwal, L.~Ahmad, and e.~a. Ilge~Akkaya, ``Gpt-4 technical report,'' 2024. [Online]. Available: \url{https://arxiv.org/abs/2303.08774}
\BIBentrySTDinterwordspacing

\bibitem{GEMINI}
\BIBentryALTinterwordspacing
G.~Team, R.~Anil, S.~Borgeaud, J.-B. Alayrac, J.~Yu, and e.~a. Radu~Soricut, ``Gemini: A family of highly capable multimodal models,'' 2024. [Online]. Available: \url{https://arxiv.org/abs/2312.11805}
\BIBentrySTDinterwordspacing

\bibitem{LLAMA}
\BIBentryALTinterwordspacing
H.~Touvron, L.~Martin, K.~Stone, P.~Albert, A.~Almahairi, and e.~a. Yasmine~Babaei, ``Llama 2: Open foundation and fine-tuned chat models,'' 2023. [Online]. Available: \url{https://arxiv.org/abs/2307.09288}
\BIBentrySTDinterwordspacing

\bibitem{STARCODER}
\BIBentryALTinterwordspacing
R.~Li, L.~B. Allal, Y.~Zi, N.~Muennighoff, D.~Kocetkov, and e.~a. Chenghao~Mou, ``Starcoder: may the source be with you!'' 2023. [Online]. Available: \url{https://arxiv.org/abs/2305.06161}
\BIBentrySTDinterwordspacing

\bibitem{C3}
\BIBentryALTinterwordspacing
X.~Dong, C.~Zhang, Y.~Ge, Y.~Mao, Y.~Gao, lu~Chen, J.~Lin, and D.~Lou, ``C3: Zero-shot text-to-sql with chatgpt,'' 2023. [Online]. Available: \url{https://arxiv.org/abs/2307.07306}
\BIBentrySTDinterwordspacing

\bibitem{DIN}
M.~Pourreza and D.~Rafiei, ``Din-sql: Decomposed in-context learning of text-to-sql with self-correction,'' in \emph{ADVANCES IN NEURAL INFORMATION PROCESSING SYSTEMS 36 (NEURIPS 2023)}, ser. Advances in Neural Information Processing Systems, A.~Oh, T.~Neumann, A.~Globerson, K.~Saenko, M.~Hardt, and S.~Levine, Eds., 2023, 37th Conference on Neural Information Processing Systems (NeurIPS), New Orleans, LA, DEC 10-16, 2023.

\bibitem{DAIL}
D.~Gao, H.~Wang, Y.~Li, X.~Sun, Y.~Qian, B.~Ding, and J.~Zhou, ``\BIBforeignlanguage{English}{Text-to-sql empowered by large language models: A benchmark evaluation},'' \emph{\BIBforeignlanguage{English}{PROCEEDINGS OF THE VLDB ENDOWMENT}}, vol.~17, no.~5, pp. 1132--1145, JAN 2024.

\bibitem{SELF-CONSISTENCY}
\BIBentryALTinterwordspacing
X.~Wang, J.~Wei, D.~Schuurmans, Q.~V. Le, E.~H. Chi, S.~Narang, A.~Chowdhery, and D.~Zhou, ``Self-consistency improves chain of thought reasoning in language models,'' in \emph{ICLR 2023}, 2023. [Online]. Available: \url{https://arxiv.org/abs/2203.11171}
\BIBentrySTDinterwordspacing

\bibitem{MSC}
\BIBentryALTinterwordspacing
S.~K. Gorti, I.~Gofman, Z.~Liu, J.~Wu, N.~Vouitsis, G.~Yu, J.~C. Cresswell, and R.~Hosseinzadeh, ``Msc-sql: Multi-sample critiquing small language models for text-to-sql translation,'' 2024. [Online]. Available: \url{https://arxiv.org/abs/2410.12916}
\BIBentrySTDinterwordspacing

\bibitem{SELF-POLISH}
\BIBentryALTinterwordspacing
Z.~Xi, S.~Jin, Y.~Zhou, R.~Zheng, S.~Gao, J.~Liu, T.~Gui, Q.~Zhang, and X.~Huang, ``Self-{P}olish: Enhance reasoning in large language models via problem refinement,'' in \emph{Findings of the Association for Computational Linguistics: EMNLP 2023}, H.~Bouamor, J.~Pino, and K.~Bali, Eds.\hskip 1em plus 0.5em minus 0.4em\relax Singapore: Association for Computational Linguistics, Dec. 2023, pp. 11\,383--11\,406. [Online]. Available: \url{https://aclanthology.org/2023.findings-emnlp.762/}
\BIBentrySTDinterwordspacing

\bibitem{DEEPSEEK-R1}
Guo, Daya, Yang, Dejian, Zhang, and e.~a. Haowei, ``Deepseek-r1: Incentivizing reasoning capability in llms via reinforcement learning,'' \emph{arXiv preprint arXiv:2501.12948}, 2025.

\bibitem{MPNET}
\BIBentryALTinterwordspacing
K.~Song, X.~Tan, T.~Qin, J.~Lu, and T.-Y. Liu, ``Mpnet: Masked and permuted pre-training for language understanding,'' in \emph{Advances in Neural Information Processing Systems}, H.~Larochelle, M.~Ranzato, R.~Hadsell, M.~Balcan, and H.~Lin, Eds., vol.~33.\hskip 1em plus 0.5em minus 0.4em\relax Curran Associates, Inc., 2020, pp. 16\,857--16\,867. [Online]. Available: \url{https://proceedings.neurips.cc/paper_files/paper/2020/file/c3a690be93aa602ee2dc0ccab5b7b67e-Paper.pdf}
\BIBentrySTDinterwordspacing

\bibitem{RESDSQL}
\BIBentryALTinterwordspacing
H.~Li, J.~Zhang, C.~Li, and H.~Chen, ``Resdsql: decoupling schema linking and skeleton parsing for text-to-sql,'' in \emph{Proceedings of the Thirty-Seventh AAAI Conference on Artificial Intelligence and Thirty-Fifth Conference on Innovative Applications of Artificial Intelligence and Thirteenth Symposium on Educational Advances in Artificial Intelligence}, ser. AAAI'23/IAAI'23/EAAI'23.\hskip 1em plus 0.5em minus 0.4em\relax AAAI Press, 2023. [Online]. Available: \url{https://doi.org/10.1609/aaai.v37i11.26535}
\BIBentrySTDinterwordspacing

\bibitem{DEEPSEEK}
\BIBentryALTinterwordspacing
DeepSeek-AI, A.~Liu, B.~Feng, B.~Xue, B.~Wang, and e.~a. Bochao~Wu, ``Deepseek-v3 technical report,'' 2024. [Online]. Available: \url{https://arxiv.org/abs/2412.19437}
\BIBentrySTDinterwordspacing

\end{thebibliography}

\end{document}